\documentclass[11pt,letterpaper]{article}
\usepackage{acl2017}
\usepackage{times}
\usepackage{latexsym}
\usepackage{epsfig}
\usepackage{graphicx}
\usepackage{amsmath}
\usepackage{amssymb}
\usepackage{graphicx}
\usepackage{tabularx, booktabs}
\usepackage{multirow}
\usepackage{subcaption}
\usepackage{comment}
\usepackage{url}
\usepackage{array}
\usepackage[utf8x]{inputenc}
\newcolumntype{P}[1]{>{\centering\arraybackslash}p{#1}}

\aclfinalcopy 

\newcommand{\ignore}[1]{}

\title{Adversarial Generation of Natural Language}

\author{Sai Rajeswar$^{\spadesuit}$\thanks{Indicates first authors. Ordering determined by coin flip.} ~ Sandeep Subramanian$^{\spadesuit}$\footnotemark[1] ~ Francis Dutil$^{\spadesuit}$ \\ ~ \textbf{Christopher Pal $^{\clubsuit\spadesuit}$ ~ Aaron Courville{$^\spadesuit\dagger$}}\\
   $^\spadesuit$MILA, Université de Montréal ~~ $^\clubsuit$École Polytechnique de Montréal ~~$^\dagger$CIFAR Fellow \\
   { \tt $\{$sai.rajeswar.mudumba,sandeep.subramanian.1,aaron.courville\}@umontreal.ca, } \\ { \tt  frdutil@gmail.com, christopher.pal@polymtl.ca
}
}

\begin{document}

\maketitle

\begin{abstract}
  Generative Adversarial Networks (GANs) have gathered a lot of attention from the computer vision community, yielding impressive results for image generation. Advances in the adversarial generation of natural language from noise however are not commensurate with the progress made in generating images, and still lag far behind likelihood based methods. 
In this paper, we take a step towards generating natural language  with a GAN objective alone. We introduce a simple baseline that addresses the discrete output space problem without relying on gradient estimators and show that it is able to achieve state-of-the-art results on a Chinese poem generation dataset. We present quantitative results on generating sentences from context-free and probabilistic context-free grammars, and qualitative language modeling results. A conditional version is also described that can generate sequences conditioned on sentence characteristics.
\end{abstract}

\section{Introduction}
Deep neural networks have recently enjoyed some success at modeling natural language \cite{mikolov2010recurrent, zaremba2014recurrent, kim2015character}. Typically, recurrent and convolutional language models are trained to maximize the likelihood of observing a word or character given the previous observations in the sequence $P(w_1 \ldots w_n) = p(w_1) \prod_{i=2}^{n} P(w_i|w_1 \ldots w_{i-1})$. These models are commonly trained using a technique called \textit{teacher forcing} \cite{williams1989learning} where the inputs to the network are fixed and the model is trained to predict only the next item in the sequence given all previous observations. This corresponds to maximum-likelihood training of these models. However this one-step ahead prediction during training makes the model prone to \textit{exposure bias} \cite{ranzato2015sequence, bengio2015scheduled}. Exposure bias occurs when a model is only trained conditioned on ground-truth contexts and is not exposed to its own errors \citep{Wiseman16beam}. An important consequence to exposure bias is that generated sequences can degenerate as small errors accumulate.  
Many important problems in NLP such as machine translation and abstractive summarization are trained via a maximum-likelihood training objective \cite{bahdanau2014neural, rush2015neural}, but require the generation of extended sequences and are evaluated based on sequence-level metrics such as BLEU \cite{papineni2002bleu} and ROUGE \cite{lin2004rouge}. 

One possible direction towards incorporating a sequence-level training objective is to use Generative Adversarial Networks (GANs) \cite{goodfellow2014generative}. While GANs have yielded impressive results for modeling images \cite{radford2015unsupervised, dumoulin2016adversarially}, advances in their use for natural language generation has lagged behind. 
Some progress has been made recently in incorporating a GAN objective in sequence modeling problems including natural language generation. 
\citet{lamb2016professor} use an adversarial criterion to match the hidden state dynamics of a teacher forced recurrent neural network (RNN) and one that samples from its own output distribution across multiple time steps. Unlike the approach in \citet{lamb2016professor}, sequence GANs \cite{yu2016seqgan} and maximum-likelihood augmented GANs \cite{che2017maximum} use an adversarial loss at outputs of an RNN. Using a GAN at the outputs of an RNN however isn't trivial since sampling from these outputs to feed to the discriminator is a non-differentiable operation. As a result gradients cannot propagate to the generator from the discriminator. \citet{yu2016seqgan} use policy gradient to estimate the generator's gradient and \cite{che2017maximum} present an importance sampling based technique. Other alternatives include REINFORCE \cite{williams1992simple}, the use of a Gumbel softmax \cite{jang2016categorical} and the straighthrough estimator \cite{bengio2013estimating} among others.

In this work, we address the discrete output space problem by simply forcing the discriminator to operate on continuous valued output distributions. The discriminator sees a sequence of probabilities over every token in the vocabulary from the generator and a sequence of 1-hot vectors from the true data distribution as in Fig. \ref{cnn_arch}. This technique is identical to that proposed by \citet{gulrajani2017improved}, which is parallel work to this. In this paper we provide a more complete empirical investigation of this approach to applying GANs to discrete output spaces. We present results using recurrent as well as convolutional architectures on three language modeling datasets of different sizes at the word and character-level. We also present quantitative results on generating sentences that adhere to a simple context-free grammar (CFG), and a richer probabilistic context-free grammar (PCFG). We compare our method to previous works that use a GAN objective to generate natural language, on a Chinese poetry generation dataset. In addition, we present a conditional GAN \cite{mirza2014conditional} that generates sentences conditioned on sentiment and questions.

\section{Generative Adversarial Networks}
GANs \cite{goodfellow2014generative} are a general framework used in training generative models by formulating the learning process as a two player minimax game as formulated in the equation below. A generator network G tries to generate samples that are as close as possible to the true data distribution $P(x)$ of interest from a fixed noise distribution $P(z)$. We will refer to the samples produced by the generator as $G(z)$. A discriminator network is then trained to distinguish between $G(z)$ and samples from the true data distribution $P(x)$ while the generator network is trained using gradient signals sent by the discriminator by minimizing $\log(1 - D(G(z)))$. 
\citet{goodfellow2014generative} have shown that, with respect to an optimal discriminator, the minimax formulation can be shown to minimize the Jensen Shannon Divergence (JSD) between the generator's output distribution and the true data distribution.
\begin{align*}
\displaystyle \min_{G} \displaystyle \max_{D} V(D, G) = \mathop{\mathbb{E}}_{x \sim P(x)} [\log D(x)] \\  + \mathop{\mathbb{E}}_{z \sim P(z)} [\log(1 - D(G(z)))]
\end{align*}
However, in practice, the generator is trained to maximize $\log(D(G(z)))$ instead, since it provides stronger gradients in the early stages of learning \cite{goodfellow2014generative}.

GANs have been reported to be notoriously hard to train in practice \cite{arjovsky2017towards} and several techniques have been proposed to alleviate some of the complexities involved in getting them to work including modified objective functions and regularization \cite{salimans2016improved, arjovsky2017wasserstein, mao2016least,gulrajani2017improved}. We discuss some of these problems in the following subsection.

\citet{nowozin2016f} show that it is possible to train GANs with a variety of f-divergence measures besides JSD. Wasserstein GANs (WGANs) \cite{arjovsky2017wasserstein} minimize the earth mover's distance or Wasserstein distance, while Least Squared GANs (LSGANs) \cite{mao2016least} modifies replaces the log loss with an L2 loss. WGAN-GP \cite{gulrajani2017improved} incorporate a gradient penalty term on the discriminator's loss in the WGAN objective which acts as a regularizer.
In this work, we will compare some of these objectives in the context of natural language generation.
\subsection{Importance of Wasserstein GANs}
\citet{arjovsky2017towards} argue that part of the problem in training regular GANs is that it seeks to minimize the JSD between the $G(z)$ and $P(x)$. When the generator is trying to optimized $log(1 - D(G(z)))$, the gradients that it receives vanish as the discriminator is trained to optimality. The authors also show that when trying to optimize the more practical alternative, $-log(D(G(z)))$, the generator might not suffer from vanishing gradients but receives unstable training signals. It is also important to consider the fact that highly structured data like images and language lie in low-dimensional manifolds (as is evident by studying their principal components). Wassterstein GANs \cite{arjovsky2017wasserstein} overcome some of the problems in regular GAN training by providing a softer metric to compare the distributions lying in low dimensional manifolds. A key contribution of this work was identifying the importance of a lipschitz constraint which is achieved by clamping the weights of the discriminator to lie in a fixed interval. The lipschitz constraint and training the discriminator multiple times for every generator gradient update creates a strong learning signal for the generator.

\citet{gulrajani2017improved} present an alternative to weight clamping that they call a gradient penalty to enforce lipschitzness since model performance was reported to be highly sensitive to the clamping hyperparameters. They add the following penalty to the discriminator training objective - $(||\triangledown_{G(z)} D(G(z))||_{2} - 1)^2$.
A potential concern regarding our strategy to train our discriminator to distinguish between sequence of 1-hot vectors from the true data distribution and a sequence of probabilities from the generator is that the discriminator can easily exploit the sparsity in the 1-hot vectors to reach optimality. However, Wassterstein distance with a lipschitz constraint / gradient penalty provides good gradients even under an optimal discriminator and so isn't a problem for us in practice. Even though it is possible to extract some performance from a regular GAN objective with the gradient penalty (as we show in one of our experiments), WGANs still provide better gradients to the generator since the discriminator doesn't saturate often.


\section{Model architecture}
\begin{figure}
 \begin{center}
\hspace{0.7cm}
\includegraphics[width=8cm,height=6cm,keepaspectratio]{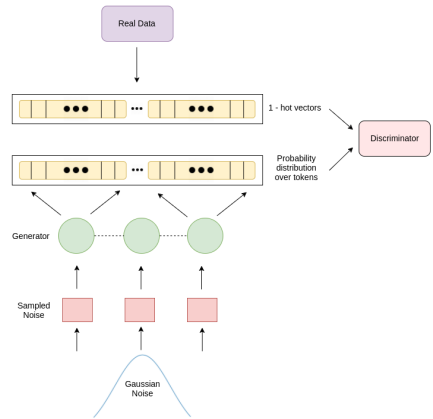}
\end{center}
   \caption{Model architecture}
\label{cnn_arch}
\end{figure}

Let $\textbf{z} \sim \mathcal{N}(\mathbf{0}, \mathbf{I})$ be the input to our generator network $G$ from which we will attempt to generate natural language. For implementation convenience, the sample $\mathbf{z}$ is of shape $n \times d$ where $n$ is the length of sequence and $d$ is a fixed length dimension of the noise vector at each time step. The generator then transforms $\mathbf{z}$ into a sequence of probability distributions over the vocabulary $G(z)$ of size $n \times k$ where $k$ is the size of our true data distribution's vocabulary. The discriminator network $D$ is provided with fake samples $G(z)$ and samples from the true data distribution $P(x)$. Samples from the true distribution are provided as a sequence of 1-hot vectors with each vector serving as an indicator of the observed word in the sample. As described in section 2, the discriminator is trained to discriminate between real and fake samples and the generator is trained to fool the discriminator as in Fig. \ref{cnn_arch}.

We investigate recurrent architectures as in \cite{lamb2016professor,yu2016seqgan,che2017maximum} and convolutional architectures in both the generator as well as the discriminator. The following subsections detail our architectures.
\subsection{Recurrent Models}
Recurrent Neural Networks (RNNs), particularly Long short-term memory networks (LSTMs) \cite{hochreiter1997long} and Gated Recurrent Networks \cite{cho2014learning} are powerful models that have been successful at modeling sequential data \cite{graves2009offline,mikolov2010recurrent}. They transform a sequence of input vectors $ \mathbf{x} = x_1 \ldots x_n$ into a sequence of hidden states $\mathbf{h} = h_1 \ldots h_n$ where each hidden state maintains a summary of the input up until then. RNN language models are autoregressive in nature since the input to the network at time $t$ depends on the output at time $t-1$. However, in the context of generating sequences from noise, the inputs are pre-determined and there is no direct correspondence between the output at time $t-1$ and the input at time $t$ this fundamentally changes the auto-regressiveness of the RNN. The RNN does however carry forward information about its output at time $t$ through subsequent time steps via its hidden states $\mathbf{h}$ as evident from its recurrent transition function. In order to incorporate an explicit dependence between subsequent RNN outputs, we add a peephole connection between the \textit{output} probability distribution $\mathbf{y_{t-1}}$ at time $t-1$ and the hidden state $\mathbf{h_t}$ at time $t$ as show in the LSTM equations below.
 Typical RNN language models have a shared affine transformation matrix $\mathbf{W_{out}}$ that is shared across time all steps that projects the hidden state vector to a vector of the same size as the target vocabulary to generate a sequence of outputs $\mathbf{y} = y_1 \ldots y_t$. Subsequently a softmax function is applied to each vector to turn it into a probability distribution over the vocabulary.
\begin{align*}
\mathbf{y}_{t} &= \mathrm{softmax}(\mathbf{W}_{out}\mathbf{h}_{t} + \mathbf{b}_{out}),
\end{align*}
During inference, an output is sampled from the softmax distribution and becomes the input at the subsequent time step. While training the inputs are pre-determined. In all of our models, we perform greedy decoding where we always pick $\operatorname{argmax} y_t$. When using the LSTM as a discriminator we use a simple binary logistic regression layer on the last hidden state $h_{n}$ to determine the probability of the sample being from the generator's data distribution or from the real data distribution. $P(real) = \sigma(\mathbf{W}_{pred}\mathbf{h}_{n} + \mathbf{b}_{pred})$.

The LSTM update equations with an output peephole are :
\begin{align*}
\mathbf{i}_{t} &= \sigma(\mathbf{W}_{xi}\mathbf{x}_{t} + \mathbf{W}_{hi}\mathbf{h}_{t-1} + \mathbf{W}_{pi}\mathbf{y}_{t-1} + \mathbf{b}_{i})\\
\mathbf{f}_{t} &= \sigma(\mathbf{W}_{xf}\mathbf{x}_{t} + \mathbf{W}_{hf}\mathbf{h}_{t-1} + \mathbf{W}_{pf}\mathbf{y}_{t-1} + \mathbf{b}_{f})\\
\mathbf{o}_{t} &= \sigma(\mathbf{W}_{xo}\mathbf{x}_{t} + \mathbf{W}_{ho}\mathbf{h}_{t-1} + \mathbf{W}_{po}\mathbf{y}_{t-1} + \mathbf{b}_{o})\\
\mathbf{c}_{t} &= \tanh(\mathbf{W}_{xc}\mathbf{x}_{t} + \mathbf{W}_{hc}\mathbf{h}_{t-1} + \mathbf{W}_{pc}\mathbf{y}_{t-1} + \mathbf{b}_{c})\\
\mathbf{c}_{t} &= \mathbf{f}_{t} \odot \mathbf{c}_{t-1} + \mathbf{i}_{t} \odot \mathbf{c}_{t}\\
\mathbf{h}_{t} &= \mathbf{o}_{t}\odot\tanh(\mathbf{c}_{t}),
\end{align*}
%
where $\sigma$ is the element-wise sigmoid function, $\odot$ is the hadamard product, $\tanh$ is the element-wise $\tanh$ function. $\mathbf{W_{\cdot}}$ and $\mathbf{b_{\cdot}}$ are learn-able parameters of the model and $\mathbf{i_t}$, $\mathbf{f_t}$, $\mathbf{o_t}$ and $\mathbf{c_t}$ constitute the input, forget, output and cell states of the LSTM respectively. 

\subsection{Convolutional Models}
Convolutional neural networks (CNNs) have also shown promise at modeling sequential data using 1-dimensional convolutions \cite{dauphin2016language, zhang2015character}. Convolution filters are convolved across time and the input dimensions are treated as channels. In this work, we explore convolutional generators and discriminators with residual connections \cite{he2016deep}.

\citet{gulrajani2017improved} use a convolutional model for both the generator and discriminator. The generator consists of 5 residual blocks with 2 1-D convolutional layers each. A final 1-D convolution layer transforms the output of the residual blocks into a sequence of un-normalized vectors for each element in the input sequence (noise). These vectors are then normalized using the softmax function. All convolutions are 'same' convolutions with a stride of 1 followed by batch-normalization \cite{ioffe2015batch} and the ReLU \cite{nair2010rectified, glorot2011deep} activation function without any pooling so as to preserve the shape of the input. The discriminator architecture is identical to that of the generator with the final output having a single output channel.

\subsection{Curriculum Learning}
In likelihood based training of generative language models, models are only trained to make one-step ahead predictions and as a result it is possible to train these models on relatively long sequences even in the initial stages of training. However, in our adversarial formulation, our generator is encouraged to generate entire sequences that match the true data distribution without explicit supervision at each step of the generation process. As a way to provide training signals of incremental difficulty, we use curriculum learning \cite{bengio2009curriculum} and train our generator to produce sequences of gradually increasing lengths as training progresses.

\section{Experiments \& Data}
GAN based methods have often been critiqued for lacking a concrete evaluation strategy \cite{salimans2016improved}, however recent work \cite{wu2016quantitative} uses an annealed importance based technique to overcome this problem.

In the context of generating natural language, it is possible to come up with a simpler approach to evaluate compute the likelihoods of generated samples. We synthesize a data generating distribution under which we can compute likelihoods in a tractable manner. We propose a simple evaluation strategy for evaluating adversarial methods of generating natural language by constructing a data generating distribution from a CFG or P$-$CFG. It is possible to determine if a sample belongs to the CFG or the probability of a sample under a P$-$CFG by using a constituency parser that is provided with all of the productions in a grammar. \citet{yu2016seqgan} also present a simple idea to estimate the likelihood of generated samples by using a randomly initialized LSTM as their data generating distribution. While this is a viable strategy to evaluate generative models of language, a randomly initialized LSTM provides little visibility into the complexity of the data distribution itself and presents no obvious way to increase its complexity. CFGs and PCFGs however, provide explicit control of the complexity via their productions. They can also be learned via grammar induction \cite{brill1993automatic} on large treebanks of natural language and so the data generating distribution is not synthetic as in \cite{yu2016seqgan}.

Typical language models are evaluated by measuring the likelihood of samples from the true data distribution under the model. However, with GANs it is impossible to measure likelihoods under the model itself and so we measure the likelihood of the model's samples under the true data distribution instead.

We divide our experiments into four categories:
\begin{itemize}
	\item Generating language that belongs to a toy CFG and an induced PCFG from the Penn Treebank \cite{marcus1993building}. 
	\item Chinese poetry generation with comparisons to \cite{yu2016seqgan} and \cite{che2017maximum}.
    \item Generated samples from a dataset consisting of simple English sentences, the 1-billion-word and Penn Treebank datasets.
    \item Conditional GANs that generate sentences conditioned on certain sentence attributes such as sentiment and questions.
\end{itemize}
\subsection{Simple CFG}
We use a simple and publicly available CFG\footnote{\url{http://www.cs.jhu.edu/~jason/465/hw-grammar/extra-grammars/holygrail}} that contains 248 productions. We then generate two sets of data from this CFG - one consisting of samples of length 5 and another of length 11. Each set contains 100,000 samples selected at random from the CFG. The first set has a vocabulary of 36 tokens while the second 45 tokens. We evaluate our models on this task by measuring the fraction of generated samples that satisfy the rules of the grammar and also measure the diversity in our generated samples. We do this by generating 1,280 samples from noise and computing the fraction of those that are valid under our grammar using the Earley parsing algorithm \cite{earley1970efficient}. In order to measure sample diversity, we simply the count the number of unique samples; while this assumes that all samples are orthogonal it still serves as a proxy measure of the entropy. We compare various generator, discriminator and GAN objectives on this problem.
\subsection{Penn Treebank PCFG}
To construct a more challenging problem than a simple CFG, we use sections 0-21 of the WSJ subsection of the Penn Treebank to induce a PCFG using simple count statistics of all productions.
\begin{align*}
P(A \rightarrow B C) = \dfrac{count(A \rightarrow B C)}{count(A \rightarrow *)}
\end{align*}
We train our model on all sentences in the treebank and restrict the output vocabulary to the top 2,000 most frequently occurring words. We evaluate our models on this task by measuring the likelihood of a sample using a Viterbi chart parser \cite{klein2003parsing}. While such a measure mostly captures the grammaticality of a sentence, it is still a reasonable proxy of sample quality.
\subsection{Chinese Poetry}
\citet{zhang2014chinese} present a dataset of Chinese poems that were used to evaluate adversarial training methods for natural language in \cite{yu2016seqgan} and \cite{che2017maximum}. The dataset consists of 4-line poems with a variable number of characters in each line. We treat each line in a poem as a training example and use lines of length 5 (poem-5) and 7 (poem-7) with the train/validation/test split\footnote{\url{http://homepages.inf.ed.ac.uk/mlap/Data/EMNLP14/}} specified in \cite{che2017maximum}. We use BLEU-2 and BLEU-3 to measure model performance on this task. Since there is no obvious "target" for each generated sentence, both works report corpus-level BLEU measures using the entire test set as the reference.
\subsection{Language Generation}
We generate language from three different datasets of varying sizes and complexity. A dataset comprising simple English sentences\footnote{\url{https://github.com/clab/sp2016.11-731/tree/master/hw4/data}} which we will henceforth refer to as CMU$-$SE, the version of the Penn Treebank commonly used in language modeling experiments \cite{zaremba2014recurrent} and the Google 1-billion word dataset \cite{chelba2013one}. We perform experiments at generating language at the word as well as character-level. The CMU$-$SE dataset consists of 44,016 sentences with a vocabulary of 3,122 words, while the Penn Treebank consists of 42,068 sentences with a vocabulary of 10,000 words. We use a random subset of 3 million sentences from the 1-billion word dataset and constrain our vocabulary to the top 30,000 most frequently occurring words. We use a curriculum learning strategy in all of our LSTM models (with and without the output peephole connection) that starts training on sentences of length 5 at the word level and 13 for characters and increases the sequence length by 1 after a fixed number of epochs based on the size of the data. Convolutional methods in \cite{gulrajani2017improved} are able to generate long sequences even without a curriculum, however we found it was critical in generating long sequences with an LSTM.

\begin{figure}[h!]
 \begin{center}
\includegraphics[width=6cm,height=4cm]{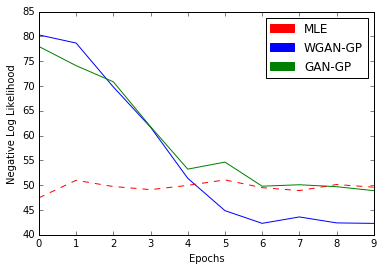}
\end{center}
   \caption{Negative log-likelihood of generated samples under the PCFG using an LSTM trained with the WGAN-GP, GAN-GP and a standard MLE objective on the PTB dataset}
\label{pcfg}
\end{figure}

\begin{table*}[h!]
\begin{center}
\begin{tabular}{| c| c| c| c| c| c| c| c| c|}
\hline
Gen & Disc & Objective & \multicolumn{2}{c|}{Length 5} & \multicolumn{2}{c|}{Length 11}\\
\hline
& & & Acc (\%) & Uniq & Acc (\%) & Uniq  \\
\hline
LSTM & LSTM & GAN & 99.06 & 0.913 & 0 & 0.855\\
\hline 
LSTM & LSTM & LSGAN & 99.45 & 0.520 & 0 & 0.855\\
\hline 
LSTM & LSTM & WGAN  & 93.98  & 0.972 & 98.04
 & 0.924\\
\hline
LSTM-P & LSTM & WGAN & 97.96 & 0.861 & 99.29
 & 0.653\\
\hline
LSTM & LSTM & WGAN-GP & 99.21 & 0.996 & 96.25 & 0.992\\
\hline
CNN & CNN & WGAN-GP & 98.59	& 0.990 & 97.01 & 0.771\\
\hline
LSTM-P & LSTM & GAN-GP & 98.68 & 0.993 & 96.32 & 0.995\\
\hline
\end{tabular}
\end{center}
\caption {Accuracy and uniqueness measure of samples generated by different models. LSTM, LSTM-P refers to the LSTM model with the output peephole and the WGAN-GP and GAN-GP refer to models that use a gradient penalty in the discriminator's training objective}
\label{cfg}
\end{table*}
\subsection{Conditional Generation of Sequences}
GANs are able to leverage explicit conditioning on high-level attributes of data \cite{mirza2014conditional, gauthier2014conditional,radford2015unsupervised} to generate samples which contain these attributes. Recent work \cite{hu2017controllable} generates sentences conditioned on certain attributes of language such as sentiment using a variational autoencoders (VAEs) \cite{kingma2013auto} and  holistic attribute discriminators. In this paper, we use two features inherent in language - sentiment and questions. To generate sentences that are questions, we use the CMU$-$SE dataset and label sentences that contain a "?" as being questions and the rest as been statements. To generate sentences of positive and negative sentiment we use the Amazon review polarity dataset collected in \cite{zhang2015character} and use the first 3 million \textit{short} reviews with a vocabulary of the top 4,000 most frequently occurring words. Conditioning on sentence attributes is achieved by concatenating a single feature map containing either entirely ones or zeros to indicate the presence or absence of the attribute as in \cite{radford2015unsupervised} at the output of each convolutional layer. The conditioning is done on both the generator and the discriminator. We experiment with conditional GANs using only convolutional methods since methods adding conditioning information has been well studied in these architectures.

\begin{table*}[htb!] 
\begin{center}
\begin{tabular}{| c| c| c| c| c| c| c| c| c|}
\hline
Models & \multicolumn{4}{c|}{Poem 5} & \multicolumn{4}{c|}{Poem 7}\\
\hline
& \multicolumn{2}{c|}{BLEU-2} & \multicolumn{2}{c|}{BLEU-3} & \multicolumn{2}{c|}{BLEU-2} & \multicolumn{2}{c|}{BLEU-3} \\
\hline
& Val & Test & Val & Test & Val & Test & Val & Test \\
\hline
MLE \cite{che2017maximum} & - & 0.693 & - & - & - & 0.318 & - & -\\
\hline 
Sequence GAN \cite{yu2016seqgan} & - & 0.738 & - & - & - & - & - & -\\
\hline
MaliGAN-basic \cite{che2017maximum} & - & 0.740 & - & - & - & 0.489 & - & -\\
\hline
MaliGAN-full \cite{che2017maximum} & - & 0.762 & - & - & - & 0.552 & - & -\\
\hline
LSTM (ours) & 0.840 & 0.837 & 0.427 & \textbf{0.372} & 0.660 & 0.655 & 0.386 & \textbf{0.405}\\
\hline
LSTM Peephole (ours) & 0.845 & \textbf{0.878} & 0.439 & 0.363 & 0.670 & \textbf{0.670} & 0.327 & 0.355\\
\hline 
\end{tabular}
\end{center}
\caption {BLEU scores on the poem-5 and poem-7 datasets}
\label{poem_generation}
\end{table*}

\subsection{Training}
All models are trained using the back-propagation algorithm updating our
parameters using the Adam optimization method \cite{kingma2014adam} and stochastic gradient descent (SGD) with batch sizes of 64. A learning rate of $2 \times 10^{-3}$, $\beta_1 = 0.5$ and $\beta_2 = 0.999$ is used in our LSTM generator and discriminators while convolutional architectures use a learning rate of $1 \times 10^{-4}$. The noise prior and all LSTM hidden dimensions are set to 128 except for the Chinese poetry generation task where we set it to 64.

\section{Results and Discussion}
Table. \ref{cfg} presents quantitative results on generating sentences that adhere to the simple CFG described in Section 4.1. The Acc column computes the accuracy with which our model generates samples from the CFG using a sample of 1,280 generations. We observe that all models are able to fit sequences of length 5 but only the WGAN, WGAN-GP objectives are able to generalize to longer sequences of length 11. This motivated us to use only the WGAN and WGAN-GP objectives in our subsequent experiments. The GAN-GP criterion appears to perform reasonably as well but we restrict our experiments to use the WGAN and WGAN-GP criteria only. GANs have been shown to exhibit the phenomenon of "mode dropping" where the generator fails to capture a large fraction of the modes present in the data generating distribution \cite{che2016mode}. It is therefore important to study the diversity in our generated samples. The Uniq column computes the number of unique samples in a sample 1,280 generations serves as a rough indicator of sample diversity. The WGAN-GP objective appears to encourage the generation of diverse samples while also fitting the data distribution well.

Fig. \ref{pcfg} shows the negative-log-likelihood of generated samples using a LSTM architecture using the WGAN-GP, GAN-GP and MLE criteria. All models used an LSTM generator. The sequence length is set to 7 and the likelihoods are evaluated at the end of every epoch on a set of 64 samples.

Table. \ref{poem_generation} contains quantitative results on the Chinese poetry generation dataset. The results indicate that our straightforward strategy to overcome back-propagating through discrete states is competitive and outperforms more complicated methods.

Table. \ref{cond_gen} contains sequences generated by our model conditioned on sentiment (positive/negative) and questions/statements. The model is able to pick up on certain consistent patterns in questions as well as when expressing sentiment and use them while generating sentences.

Tables  \ref{1_billion} and \ref{ptb_cmu} contain sequences generated at the word and character-level by our LSTM and CNN models. Both models are able to produce realistic sentences. The CNN model with a WGAN-GP objective appears to be able to maintain context over longer time spans.
\begin{table*}[htb!]
\begin{center}
\begin{tabular}{|p{2cm}|p{2cm}|p{11cm}|}
\hline
Level & Method & 1-billion-word \\
\hline
\multirow{12}{*}{Word}
      & \multirow{6}{*}{LSTM} &An opposition was growing in China . \\
      & & This is undergoing operation a year . \\
      & & It has his everyone on a blame . \\
      &  &Everyone shares that Miller seems converted President  as Democrat .\\
      &  & Which is actually the best of his children . \\
      & & Who has The eventual policy and weak ? 
\\\cline{2-3}
      & \multirow{4}{*}{CNN} & Companies I upheld , respectively patented saga and Ambac. \\ 
      & & Independence Unit have any will MRI in these Lights \\ 
      & & It is a wrap for the annually of  Morocco \\
      & & The town has Registration matched with unk and the citizens \\
\hline
\multirow{4}{*}{Character} 
      & \multirow{4}{*}{CNN} & To holl is now my Hubby ,\\
      & & The gry timers was faller\\
      & & After they work is jith a\\
      & & But in a linter a revent\\
\hline
\end{tabular}
\end{center}
\caption{Word and character-level generations on the 1-billion word dataset}
\label{1_billion}
\end{table*}

\begin{table*}[htb!]
\begin{center}
\begin{tabular}{|p{0.8cm}|p{0.9cm}|p{6.5cm}|p{6.8cm}|}
\hline
Level & Model & PTB & CMU-SE \\
\hline
\multirow{12}{*}{Word}
      & \multirow{7}{*}{LSTM} & what everything they take everything away from . & \textless s\textgreater will you have two moment ? \textless/s\textgreater \\
      & & may tea bill is the best chocolate from emergency . &  \textless s\textgreater i need to understand deposit length . \textless/s\textgreater \\ 
      & & can you show show if any fish left inside . & \textless s\textgreater how is the another headache ?  \textless/s\textgreater \\
      & &room service , have my dinner please . & \textless s\textgreater how there , is the restaurant popular this cheese ?  \textless/s\textgreater
\\\cline{2-4}
      & \multirow{4}{*}{CNN} & meanwhile henderson said that it has to bounce for. & \textless s\textgreater i 'd like to fax a newspaper . \textless/s\textgreater \\
      & & I'm at the missouri burning the indexing manufacturing and through . & \textless s\textgreater cruise pay the next in my replacement . \textless/s\textgreater \\
      & & & \textless s\textgreater what 's in the friday  food ?  ? \textless/s\textgreater \\
\hline
\end{tabular}
\end{center}
\caption{Word level generations on the Penn Treebank and CMU-SE datasets}
\label{ptb_cmu}
\end{table*}

\begin{table*}[htb!]
\begin{center}
\begin{tabular}{|p{7.5cm}|p{6.8cm}|}
\hline
  POSITIVE & NEGATIVE \\
\hline
      best and top notch newtonmom .&  usuall the review omnium nothing non-functionable \\
      good buy homeostasis money well spent & \\

      kickass cosamin of time and fun . &  extreme crap-not working and eeeeeew  \\ 
       great britani  !  I lovethis. & a horrible poor imposing se400  \\
\cline{1-2}
\hline
  QUESTION &STATEMENT \\
\hline
        \textless s\textgreater when 's the friday convention on ? \textless /s\textgreater &   \textless s\textgreater i report my run on one mineral .  \textless /s\textgreater \\
      \textless s\textgreater how many snatched crew you have ? \textless /s\textgreater& \textless s\textgreater we have to record this now . \textless /s\textgreater
 \\
      \textless s\textgreater how can you open this hall ? \textless /s\textgreater& \textless s\textgreater i think i deeply take your passenger .\textless /s\textgreater\\
\hline
\end{tabular}
\end{center}
\caption{Coditional generation of text. Top row shows generated samples conditionally trained on amazon review polarity dataset with two attributes 'positive' and 'negative'. Bottom row has samples conditioned on the 'question' attribute}
\label{cond_gen}
\end{table*}

\section{Conclusion and Future work}
In conclusion, this work presents a straightforward but effective method to train GANs for natural language. The simplicity lies in \textit{forcing the discriminator to operate on continuous values} by presenting it with a sequence of probability distributions from the generator and a sequence of 1-hot vectors corresponding to data from the true distribution. We propose an evaluation strategy that involves learning the data distribution defined by a CFG or PCFG. This lets us evaluate the likelihood of a sample belonging to the data generating distribution. The use of WGAN and WGAN-GP objectives produce realistic sentences on datasets of varying complexity (CMU-SE, Penn Treebank and the 1-billion dataset). We also show that it is possible to perform conditional generation of text on high-level sentence features such as sentiment and questions.
In future work, we would like to explore GANs in other domains of NLP such as non goal-oriented dialog systems where a clear training and evaluation criterion does not exist.
\section*{Acknowledgements}
The Authors would like to thank Ishaan Gulrajani, Martin Arjovsky, Guillaume Lample, Rosemary Ke, Juneki Hong and Varsha Embar for their advice and insightful comments. We are grateful to the Fonds de Recherche du Québec -- Nature et Technologie for their financial support. We would also like to acknowledge NVIDIA for donating a DGX-1 computer used in this work.

\section*{Appendix}
\begin{figure}[h!]
 \begin{center}
\includegraphics[width=6cm,height=6cm,keepaspectratio]
{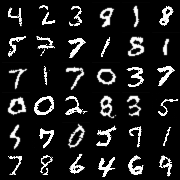}
\end{center}
   \caption{Binarized MNIST samples using a DCWGAN with gradient penalty}
\label{binary_mnist}
\end{figure}

We demonstrate that our approach to solve the problem of discrete outputs produces reasonable outputs even when applied to images. Figure \ref{binary_mnist} shows samples generated on the binarized MNIST dataset \cite{salakhutdinov2008quantitative}. We used a generator and discriminator architecture identical to \cite{radford2015unsupervised} with the WGAN-GP criterion. The generator's outputs are continuous while samples from the true data distribution are binarized.

\clearpage

\bibliography{acl2017}

\begin{thebibliography}{}
\expandafter\ifx\csname natexlab\endcsname\relax\def\natexlab#1{#1}\fi

\bibitem[{Arjovsky and Bottou(2017)}]{arjovsky2017towards}
Martin Arjovsky and L{\'e}on Bottou. 2017.
\newblock Towards principled methods for training generative adversarial
  networks.
\newblock In {\em NIPS 2016 Workshop on Adversarial Training. In review for
  ICLR\/}. volume 2016.

\bibitem[{Arjovsky et~al.(2017)Arjovsky, Chintala, and
  Bottou}]{arjovsky2017wasserstein}
Martin Arjovsky, Soumith Chintala, and L{\'e}on Bottou. 2017.
\newblock Wasserstein gan.
\newblock {\em arXiv preprint arXiv:1701.07875\/} .

\bibitem[{Bahdanau et~al.(2014)Bahdanau, Cho, and Bengio}]{bahdanau2014neural}
Dzmitry Bahdanau, Kyunghyun Cho, and Yoshua Bengio. 2014.
\newblock Neural machine translation by jointly learning to align and
  translate.
\newblock {\em arXiv preprint arXiv:1409.0473\/} .

\bibitem[{Bengio et~al.(2015)Bengio, Vinyals, Jaitly, and
  Shazeer}]{bengio2015scheduled}
Samy Bengio, Oriol Vinyals, Navdeep Jaitly, and Noam Shazeer. 2015.
\newblock Scheduled sampling for sequence prediction with recurrent neural
  networks.
\newblock In {\em Advances in Neural Information Processing Systems\/}. pages
  1171--1179.

\bibitem[{Bengio et~al.(2013)Bengio, L{\'e}onard, and
  Courville}]{bengio2013estimating}
Yoshua Bengio, Nicholas L{\'e}onard, and Aaron Courville. 2013.
\newblock Estimating or propagating gradients through stochastic neurons for
  conditional computation.
\newblock {\em arXiv preprint arXiv:1308.3432\/} .

\bibitem[{Bengio et~al.(2009)Bengio, Louradour, Collobert, and
  Weston}]{bengio2009curriculum}
Yoshua Bengio, J{\'e}r{\^o}me Louradour, Ronan Collobert, and Jason Weston.
  2009.
\newblock Curriculum learning.
\newblock In {\em Proceedings of the 26th annual international conference on
  machine learning\/}. ACM, pages 41--48.

\bibitem[{Brill(1993)}]{brill1993automatic}
Eric Brill. 1993.
\newblock Automatic grammar induction and parsing free text: A
  transformation-based approach.
\newblock In {\em Proceedings of the workshop on Human Language Technology\/}.
  Association for Computational Linguistics, pages 237--242.

\bibitem[{Che et~al.(2016)Che, Li, Jacob, Bengio, and Li}]{che2016mode}
Tong Che, Yanran Li, Athul~Paul Jacob, Yoshua Bengio, and Wenjie Li. 2016.
\newblock Mode regularized generative adversarial networks.
\newblock {\em arXiv preprint arXiv:1612.02136\/} .

\bibitem[{Che et~al.(2017)Che, Li, Zhang, Hjelm, Li, Song, and
  Bengio}]{che2017maximum}
Tong Che, Yanran Li, Ruixiang Zhang, R~Devon Hjelm, Wenjie Li, Yangqiu Song,
  and Yoshua Bengio. 2017.
\newblock Maximum-likelihood augmented discrete generative adversarial
  networks.
\newblock {\em arXiv preprint arXiv:1702.07983\/} .

\bibitem[{Chelba et~al.(2013)Chelba, Mikolov, Schuster, Ge, Brants, Koehn, and
  Robinson}]{chelba2013one}
Ciprian Chelba, Tomas Mikolov, Mike Schuster, Qi~Ge, Thorsten Brants, Phillipp
  Koehn, and Tony Robinson. 2013.
\newblock One billion word benchmark for measuring progress in statistical
  language modeling.
\newblock {\em arXiv preprint arXiv:1312.3005\/} .

\bibitem[{Cho et~al.(2014)Cho, Van~Merri{\"e}nboer, Gulcehre, Bahdanau,
  Bougares, Schwenk, and Bengio}]{cho2014learning}
Kyunghyun Cho, Bart Van~Merri{\"e}nboer, Caglar Gulcehre, Dzmitry Bahdanau,
  Fethi Bougares, Holger Schwenk, and Yoshua Bengio. 2014.
\newblock Learning phrase representations using rnn encoder-decoder for
  statistical machine translation.
\newblock {\em arXiv preprint arXiv:1406.1078\/} .

\bibitem[{Dauphin et~al.(2016)Dauphin, Fan, Auli, and
  Grangier}]{dauphin2016language}
Yann~N Dauphin, Angela Fan, Michael Auli, and David Grangier. 2016.
\newblock Language modeling with gated convolutional networks.
\newblock {\em arXiv preprint arXiv:1612.08083\/} .

\bibitem[{Dumoulin et~al.(2016)Dumoulin, Belghazi, Poole, Lamb, Arjovsky,
  Mastropietro, and Courville}]{dumoulin2016adversarially}
Vincent Dumoulin, Ishmael Belghazi, Ben Poole, Alex Lamb, Martin Arjovsky,
  Olivier Mastropietro, and Aaron Courville. 2016.
\newblock Adversarially learned inference.
\newblock {\em arXiv preprint arXiv:1606.00704\/} .

\bibitem[{Earley(1970)}]{earley1970efficient}
Jay Earley. 1970.
\newblock An efficient context-free parsing algorithm.
\newblock {\em Communications of the ACM\/} 13(2):94--102.

\bibitem[{Gauthier(2014)}]{gauthier2014conditional}
Jon Gauthier. 2014.
\newblock Conditional generative adversarial nets for convolutional face
  generation.
\newblock {\em Class Project for Stanford CS231N: Convolutional Neural Networks
  for Visual Recognition, Winter semester\/} 2014(5):2.

\bibitem[{Glorot et~al.(2011)Glorot, Bordes, and Bengio}]{glorot2011deep}
Xavier Glorot, Antoine Bordes, and Yoshua Bengio. 2011.
\newblock Deep sparse rectifier neural networks.
\newblock In {\em Aistats\/}. volume~15, page 275.

\bibitem[{Goodfellow et~al.(2014)Goodfellow, Pouget-Abadie, Mirza, Xu,
  Warde-Farley, Ozair, Courville, and Bengio}]{goodfellow2014generative}
Ian Goodfellow, Jean Pouget-Abadie, Mehdi Mirza, Bing Xu, David Warde-Farley,
  Sherjil Ozair, Aaron Courville, and Yoshua Bengio. 2014.
\newblock Generative adversarial nets.
\newblock In {\em Advances in neural information processing systems\/}. pages
  2672--2680.

\bibitem[{Graves and Schmidhuber(2009)}]{graves2009offline}
Alex Graves and J{\"u}rgen Schmidhuber. 2009.
\newblock Offline handwriting recognition with multidimensional recurrent
  neural networks.
\newblock In {\em Advances in neural information processing systems\/}. pages
  545--552.

\bibitem[{Gulrajani et~al.(2017)Gulrajani, Ahmed, Arjovsky, Dumoulin, and
  Courville}]{gulrajani2017improved}
Ishaan Gulrajani, Faruk Ahmed, Martin Arjovsky, Vincent Dumoulin, and Aaron
  Courville. 2017.
\newblock Improved training of wasserstein gans.
\newblock {\em arXiv preprint arXiv:1704.00028\/} .

\bibitem[{He et~al.(2016)He, Zhang, Ren, and Sun}]{he2016deep}
Kaiming He, Xiangyu Zhang, Shaoqing Ren, and Jian Sun. 2016.
\newblock Deep residual learning for image recognition.
\newblock In {\em Proceedings of the IEEE Conference on Computer Vision and
  Pattern Recognition\/}. pages 770--778.

\bibitem[{Hochreiter and Schmidhuber(1997)}]{hochreiter1997long}
Sepp Hochreiter and J{\"u}rgen Schmidhuber. 1997.
\newblock Long short-term memory.
\newblock {\em Neural computation\/} 9(8):1735--1780.

\bibitem[{Hu et~al.(2017)Hu, Yang, Liang, Salakhutdinov, and
  Xing}]{hu2017controllable}
Zhiting Hu, Zichao Yang, Xiaodan Liang, Ruslan Salakhutdinov, and Eric~P Xing.
  2017.
\newblock Controllable text generation.
\newblock {\em arXiv preprint arXiv:1703.00955\/} .

\bibitem[{Ioffe and Szegedy(2015)}]{ioffe2015batch}
Sergey Ioffe and Christian Szegedy. 2015.
\newblock Batch normalization: Accelerating deep network training by reducing
  internal covariate shift.
\newblock {\em arXiv preprint arXiv:1502.03167\/} .

\bibitem[{Jang et~al.(2016)Jang, Gu, and Poole}]{jang2016categorical}
Eric Jang, Shixiang Gu, and Ben Poole. 2016.
\newblock Categorical reparameterization with gumbel-softmax.
\newblock {\em arXiv preprint arXiv:1611.01144\/} .

\bibitem[{Kim et~al.(2015)Kim, Jernite, Sontag, and Rush}]{kim2015character}
Yoon Kim, Yacine Jernite, David Sontag, and Alexander~M Rush. 2015.
\newblock Character-aware neural language models.
\newblock {\em arXiv preprint arXiv:1508.06615\/} .

\bibitem[{Kingma and Ba(2014)}]{kingma2014adam}
Diederik Kingma and Jimmy Ba. 2014.
\newblock Adam: A method for stochastic optimization.
\newblock {\em arXiv preprint arXiv:1412.6980\/} .

\bibitem[{Kingma and Welling(2013)}]{kingma2013auto}
Diederik~P Kingma and Max Welling. 2013.
\newblock Auto-encoding variational bayes.
\newblock {\em arXiv preprint arXiv:1312.6114\/} .

\bibitem[{Klein and Manning(2003)}]{klein2003parsing}
Dan Klein and Christopher~D Manning. 2003.
\newblock A parsing: fast exact viterbi parse selection.
\newblock In {\em Proceedings of the 2003 Conference of the North American
  Chapter of the Association for Computational Linguistics on Human Language
  Technology-Volume 1\/}. Association for Computational Linguistics, pages
  40--47.

\bibitem[{Lamb et~al.(2016)Lamb, GOYAL, Zhang, Zhang, Courville, and
  Bengio}]{lamb2016professor}
Alex~M Lamb, Anirudh Goyal ALIAS~PARTH GOYAL, Ying Zhang, Saizheng Zhang,
  Aaron~C Courville, and Yoshua Bengio. 2016.
\newblock Professor forcing: A new algorithm for training recurrent networks.
\newblock In {\em Advances In Neural Information Processing Systems\/}. pages
  4601--4609.

\bibitem[{Lin(2004)}]{lin2004rouge}
Chin-Yew Lin. 2004.
\newblock Rouge: A package for automatic evaluation of summaries.
\newblock In {\em Text summarization branches out: Proceedings of the ACL-04
  workshop\/}. Barcelona, Spain, volume~8.

\bibitem[{Mao et~al.(2016)Mao, Li, Xie, Lau, and Wang}]{mao2016least}
Xudong Mao, Qing Li, Haoran Xie, Raymond~YK Lau, and Zhen Wang. 2016.
\newblock Least squares generative adversarial networks.
\newblock {\em arXiv preprint ArXiv:1611.04076\/} .

\bibitem[{Marcus et~al.(1993)Marcus, Marcinkiewicz, and
  Santorini}]{marcus1993building}
Mitchell~P Marcus, Mary~Ann Marcinkiewicz, and Beatrice Santorini. 1993.
\newblock Building a large annotated corpus of english: The penn treebank.
\newblock {\em Computational linguistics\/} 19(2):313--330.

\bibitem[{Mikolov et~al.(2010)Mikolov, Karafi{\'a}t, Burget, Cernock{\`y}, and
  Khudanpur}]{mikolov2010recurrent}
Tomas Mikolov, Martin Karafi{\'a}t, Lukas Burget, Jan Cernock{\`y}, and Sanjeev
  Khudanpur. 2010.
\newblock Recurrent neural network based language model.
\newblock In {\em Interspeech\/}. volume~2, page~3.

\bibitem[{Mirza and Osindero(2014)}]{mirza2014conditional}
Mehdi Mirza and Simon Osindero. 2014.
\newblock Conditional generative adversarial nets.
\newblock {\em arXiv preprint arXiv:1411.1784\/} .

\bibitem[{Nair and Hinton(2010)}]{nair2010rectified}
Vinod Nair and Geoffrey~E Hinton. 2010.
\newblock Rectified linear units improve restricted boltzmann machines.
\newblock In {\em Proceedings of the 27th international conference on machine
  learning (ICML-10)\/}. pages 807--814.

\bibitem[{Nowozin et~al.(2016)Nowozin, Cseke, and Tomioka}]{nowozin2016f}
Sebastian Nowozin, Botond Cseke, and Ryota Tomioka. 2016.
\newblock f-gan: Training generative neural samplers using variational
  divergence minimization.
\newblock In {\em Advances in Neural Information Processing Systems\/}. pages
  271--279.

\bibitem[{Papineni et~al.(2002)Papineni, Roukos, Ward, and
  Zhu}]{papineni2002bleu}
Kishore Papineni, Salim Roukos, Todd Ward, and Wei-Jing Zhu. 2002.
\newblock Bleu: a method for automatic evaluation of machine translation.
\newblock In {\em Proceedings of the 40th annual meeting on association for
  computational linguistics\/}. Association for Computational Linguistics,
  pages 311--318.

\bibitem[{Radford et~al.(2015)Radford, Metz, and
  Chintala}]{radford2015unsupervised}
Alec Radford, Luke Metz, and Soumith Chintala. 2015.
\newblock Unsupervised representation learning with deep convolutional
  generative adversarial networks.
\newblock {\em arXiv preprint arXiv:1511.06434\/} .

\bibitem[{Ranzato et~al.(2015)Ranzato, Chopra, Auli, and
  Zaremba}]{ranzato2015sequence}
Marc'Aurelio Ranzato, Sumit Chopra, Michael Auli, and Wojciech Zaremba. 2015.
\newblock Sequence level training with recurrent neural networks.
\newblock {\em arXiv preprint arXiv:1511.06732\/} .

\bibitem[{Rush et~al.(2015)Rush, Chopra, and Weston}]{rush2015neural}
Alexander~M Rush, Sumit Chopra, and Jason Weston. 2015.
\newblock A neural attention model for abstractive sentence summarization.
\newblock {\em arXiv preprint arXiv:1509.00685\/} .

\bibitem[{Salakhutdinov and Murray(2008)}]{salakhutdinov2008quantitative}
Ruslan Salakhutdinov and Iain Murray. 2008.
\newblock On the quantitative analysis of deep belief networks.
\newblock In {\em Proceedings of the 25th international conference on Machine
  learning\/}. ACM, pages 872--879.

\bibitem[{Salimans et~al.(2016)Salimans, Goodfellow, Zaremba, Cheung, Radford,
  and Chen}]{salimans2016improved}
Tim Salimans, Ian Goodfellow, Wojciech Zaremba, Vicki Cheung, Alec Radford, and
  Xi~Chen. 2016.
\newblock Improved techniques for training gans.
\newblock In {\em Advances in Neural Information Processing Systems\/}. pages
  2226--2234.

\bibitem[{Williams(1992)}]{williams1992simple}
Ronald~J Williams. 1992.
\newblock Simple statistical gradient-following algorithms for connectionist
  reinforcement learning.
\newblock {\em Machine learning\/} 8(3-4):229--256.

\bibitem[{Williams and Zipser(1989)}]{williams1989learning}
Ronald~J Williams and David Zipser. 1989.
\newblock A learning algorithm for continually running fully recurrent neural
  networks.
\newblock {\em Neural computation\/} 1(2):270--280.

\bibitem[{Wiseman and Rush(2016)}]{Wiseman16beam}
Sam Wiseman and Alexander~M. Rush. 2016.
\newblock \href{http://arxiv.org/abs/1606.02960}{Sequence-to-sequence learning
  as beam-search optimization}.
\newblock {\em CoRR\/} abs/1606.02960.
\newblock
  \href{http://arxiv.org/abs/1606.02960}{http://arxiv.org/abs/1606.02960}.

\bibitem[{Wu et~al.(2016)Wu, Burda, Salakhutdinov, and
  Grosse}]{wu2016quantitative}
Yuhuai Wu, Yuri Burda, Ruslan Salakhutdinov, and Roger Grosse. 2016.
\newblock On the quantitative analysis of decoder-based generative models.
\newblock {\em arXiv preprint arXiv:1611.04273\/} .

\bibitem[{Yu et~al.(2016)Yu, Zhang, Wang, and Yu}]{yu2016seqgan}
Lantao Yu, Weinan Zhang, Jun Wang, and Yong Yu. 2016.
\newblock Seqgan: sequence generative adversarial nets with policy gradient.
\newblock {\em arXiv preprint arXiv:1609.05473\/} .

\bibitem[{Zaremba et~al.(2014)Zaremba, Sutskever, and
  Vinyals}]{zaremba2014recurrent}
Wojciech Zaremba, Ilya Sutskever, and Oriol Vinyals. 2014.
\newblock Recurrent neural network regularization.
\newblock {\em arXiv preprint arXiv:1409.2329\/} .

\bibitem[{Zhang et~al.(2015)Zhang, Zhao, and LeCun}]{zhang2015character}
Xiang Zhang, Junbo Zhao, and Yann LeCun. 2015.
\newblock Character-level convolutional networks for text classification.
\newblock In {\em Advances in neural information processing systems\/}. pages
  649--657.

\bibitem[{Zhang and Lapata(2014)}]{zhang2014chinese}
Xingxing Zhang and Mirella Lapata. 2014.
\newblock Chinese poetry generation with recurrent neural networks.
\newblock In {\em EMNLP\/}. pages 670--680.

\end{thebibliography}
\bibliographystyle{acl_natbib}

\end{document}